\documentclass{article}

\usepackage{arxiv}

\usepackage[bookmarks=false]{hyperref}
    \hypersetup{colorlinks,
      linkcolor=blue,
      citecolor=blue,
      urlcolor=blue}

\usepackage{amssymb}
\usepackage[figuresright]{rotating}
\usepackage{amsmath}
\usepackage{booktabs}       
\usepackage{makecell}
\usepackage{subfig}
\usepackage{wrapfig}
\usepackage[numbers,square]{natbib}
\usepackage[export]{adjustbox}

\newcommand{\articleTitle}[0]{Correlation Analysis between the Robustness of Sparse Neural Networks and their Random Hidden Structural Priors}

\newcommand{\articleAuthorM}[0]{Mehdi Ben Amor}
\newcommand{\articleAuthorJ}[0]{Julian Stier}
\newcommand{\articleAuthorG}[0]{Michael Granitzer}
\newcommand{\articleAuthor}[0]{\articleAuthorM \and \articleAuthorJ \and \articleAuthorG}

\renewcommand{\shorttitle}{}

\hypersetup{
pdftitle={Corr Analysis between SNNs and Their Hidden Priors},
pdfsubject={cs-ML},
pdfauthor={Mehdi Ben Amor, Julian Stier, Michael Granitzer},
pdfkeywords={Robustness, Adversarial Attacks, Sparse Neural Networks},
}
\date{} 
\author{\articleAuthor \And
	\texttt{\{mehdi.benamor,julian.stier,michael.granitzer\}@uni-passau.de} \\
	\texttt{University of Passau}} 
\begin{document}

\title{\articleTitle}
\maketitle


\begin{abstract}
Deep learning models have been shown to be vulnerable to adversarial attacks.
This perception led to analyzing deep learning models not only from the perspective of their performance measures but also their robustness to certain types of adversarial attacks.
We take another step forward in relating the architectural structure of neural networks from a graph theoretic perspective to their robustness. 
We aim to investigate any existing correlations between graph theoretic properties and robustness of Sparse Neural Networks.
Our hypothesis is, that graph theoretic properties as a prior of neural network structures is related to their robustness.
To give answer to this hypothesis, we designed an empirical study with neural network models obtained through random graphs used as sparse structural priors for the networks. We additionally investigated the evaluation of a randomly pruned fully connected network as point of reference.

We found that robustness measures are independent to initialization methods but show weak correlations with graph properties:
higher graph densities correlate with lower robustness, but higher average path lengths and average node eccentricities show negative correlations with robustness measures.
We hope to motivate further empirical and analytical research to tightening an answer to our hypothesis.
\end{abstract}



\section{Introduction}\label{sec:introduction}
Deep learning is a leading factor for various new developments in machine learning.
It describes a revival of existing techniques and new insights which enabled tremendous progress in application areas such as computer vision and audio analysis.

While deep neural networks have shown that they can achieve a high performance on representing complex and high dimensional functions, they have exhibited vulnerability to \textbf{adversarial examples} in the form of inputs intentionally designed to cause mistakes with relatively high confidence \cite{akhtar2018threat}.
In the case of image classification, those perturbations are hardly perceptible and indistinguishable to a human observer, yet they completely fool deep learning models.
Adversarial attacks introduce a threat to the precision and the effectiveness of deep learning in real-life applications.
Up until now, researchers have been testing different deep learning models' robustness to adversarial attacks and analyzing their robustness from different perspectives.
One of the missing perspectives of studying the robustness is with respect to the graph theoretic properties describing the graph underlying the neural networks.
Particularly, we are considering Sparse Neural Networks (SNNs), i.e networks in which layers are not fully connected.

SNNs could in theory be computed much faster than their dense counterparts and in domains such as large-scale image classification are in some cases known to exploit spatial dependencies of the input domain.
Besides their computational benefits they still seem to have excellent generalization performances \cite{thom2015sparse}.
Convolutional neural networks are a highly successful example that exhibit sparse structures not only to achieve faster computation time but also to achieve better generalization \cite{lecun1995convolutional}.

Goal of this work is to study the robustness of sparse structures of neural networks.
Our hypothesis is that the robustness of neural networks and specifically sparse neural networks against adversarial examples in the form of additive perturbations is related to their graph properties.
This main research goal is cleaved on two sub-questions:
(1) Are small-world graphs meaningful prior structures for robust neural networks?
(2) Are the robustness measures of sparse networks evaluated against adversarial attacks correlated to the graph properties of these models?

Our contributions comprise (1) a transformation of random graphs into sparse neural networks by making the graphs directed and acyclic, applying a layering on the nodes, and using them as a the hidden structure of the networks, (2) extensive \textbf{robustness study on sparse neural networks} with a correlation analysis of graph properties which showed that graph density, average path length, and number of edges (weights) exhibts relationships with the robustness. 

\paragraph{Overview} \autoref{sec:related-work} introduces related work. 
In \autoref{sec:sparse-models}, we introduce definitions and notations regarding the graph theoretic prespective of neural networks. 
The adversarial attacks used as robustness measures in our study are introduced in \autoref{sec:adv-robustness}.
In \autoref{sec:experimental_design}, experiments of the two case studies are presented in detail, the robustness of pruned neural networks (to adversarial noise) and the adversarial robustness of generated SNNs.
We present our results in \autoref{sec:results} and conclude with a brief discussion illustrating our correlation analysis in \autoref{sec:conclusion}.

\section{Related Work}\label{sec:related-work}
To the best of our knowledge, no one considered analysing graph properties of neural networks and their influence on adversarial robustness.
We provide related work concerning sparsity and structure of neural networks, graph theoretical work from network science and an overview of the most prominent works in adversarial attacks for measuring robustness of neural networks.

Related work on sparsity in neural networks is rising.
Intentions for sparsity ranges from computational aspects, biological plausibility, explainability, expressiveness to compression.
Finding such networks is an ongoing research topic in areas such as regularization \cite{louizos2017learning}, pruning \cite{frankle2018lottery} and neural architecture search \cite{elsken2018neural}.
Here, we focus on the model of Watts-Strogatz as an initial structure of neural networks.
This idea was previously used in works such as \cite{stier2019structural,xie2019exploring}.

Several studies have been proposed over the past decade that are similar to our work.
Goodfellow et al. \cite{goodfellow2014explaining} conducted an empirical study of adversarial instability where the hypothesis was that adversarial perturbations comes from the linear behavior of most deep neural networks.
They suggested as well that the adversarial examples generated with FGSM (Fast Gradient Sign Method) generalize over different architectures of neural networks.
They stated that these models ``often agree with each other'' on the misclassified class when classifying an adversarial example.

Alfawzi et al. \cite{fawzi2018analysis} showed in their analysis of classifiers' instability to adversarial robustness from a mathematical perspective that adversarial instability is directly related to a \textit{distinguishability} measure that captures the difficulty of the classification task, and to the low flexbility of deep nets. 
It was also highlighted in their findings that the depth of a neural network has a positive correlation with the robustness of the classifier.

Alfawzi et al. \cite{fawzi2017geometric} showed interesting correlations between the robustness of neural networks and the geometric properties of their decision boundaries.
Szegedy et al. \cite{szegedy2013intriguing} showed an intriguing weakness of deep neural networks against small perturbations of images that makes them misclassify the perturbed examples with high prediction confidence.

Adversarial attacks are split into two classes: white box and black box attacks.
While the former assume complete knowledge of the target model's internal workings (i.e parameters, architecture etc.), the latter only considers its inputs and outputs.
The perturbed images or samples are called \textit{adversarial examples} and they represent the output of the adversarial attacks.

Goodfellow et al. propose a method that is based on the gradient of the cost function called Fast Gradient Sign Method (FGSM) that exposes the "linearity" of most deep networks \cite{goodfellow2014explaining}.
Kurakin et al.\cite{kurakin2016adversarial} introduced an extension to FGSM by applying it multiple times with small step size and clipping pixel values of intermediate results after each step to ensure that they are in a certain constrained neighborhood of the original image. 
This method was referred to as Basic Iterative Method (BIM). Later on, BIM was extended to Iterative Least-likely Class Method (ILCM) by replacing the label of the image with the least likely class predicted by the classifier.

\section{Sparse Neural Networks}\label{sec:sparse-models}
Following notations of \cite{diestel2017graph}, a graph $G = (V,E)$ consists of vertices $V$ and edges $E\subseteq[V]^2$.
Edges shortly are denoted $uv$ for $(u,v)\in E(G)$.
The vertex set of a graph $G_1$ is referred with $V(G_1)$, its edges with $E(G_1)$.
A graph's number of vertices is called the order of $G$ and given as $v_G = |G| = |V(G)|$ and the number of edges $e_G = ||G|| = |E(G)|$.
The adjancency matrix $A_G = (a_{ij})_{v_G\times v_G}$ is defined by $a_{ij} := 1$ iff $v_iv_j\in E(G)$ and $a_{ij} := 0$ otherwise.
A graph is said to be sparse if its density is in the lower range of the density's codomain,  $0 < D <\frac{1}{2}$.
Density $D$ is the proportion of the number of edges $e_G$ on all possible edges $max(e_G)$ and is given for both $D_d$ and $D_u$ simply as $D_d = \frac{|E|}{max(|E|)} = \frac{|E|}{|V|\cdot(|V|-1) } = \frac{1}{2} D_u$.
Overall, we considered the number of vertices, number of edges, overall graph density, diameter, degree distribution, path length distribution, eccentricity, betweenness and closeness centralities and refer to the networkx documentation for details \footnote{For graph properties we refer to \url{https://networkx.org} or \cite{diestel2017graph}.}.

\subparagraph{Transformation to Directed Acyclic Graphs}
Given an undirected graph $G^{u}$, deriving a directed acyclic graph $G^{d}$ can achieved using different methods, by using the lower triangular adjacency matrix of $G^{u}$ and set it as the new adjacency matrix of $G^{d}$.
Diagonal entries and the upper triangular matrix are set to zero.
This transformation loses properties and especially properties specific for undirected graphs.
We use this transformation in case we do not have a directed or acyclic version of a graph available.
An investigation on the effect of this transformation is shown with the results in section \ref{sec:results}

\begin{wrapfigure}{r}{0.4\textwidth}
	\vspace{-1.5em}
	\centering
	\includegraphics[width=0.4\textwidth]{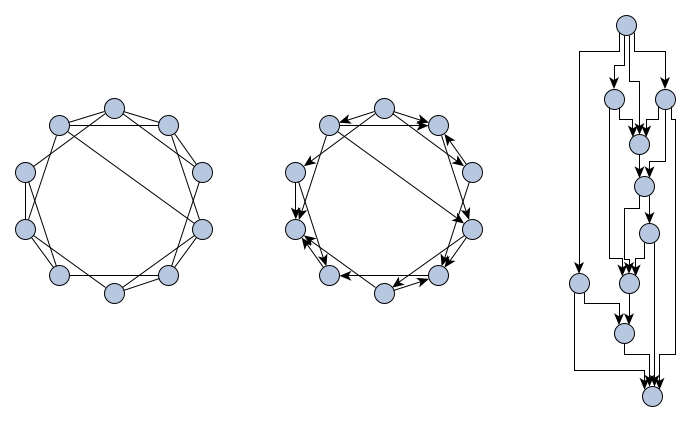}
	\caption{A randomly sampled Watts-Strogatz graph on the left, a derived directed acyclic variation of the graph as shown in the middle, and the topologically sorted prior structure on the right.}
	\label{fig:graph-induction}
\end{wrapfigure}

We investigate on the graphs of neural networks by constructing them from a Watts-Strogatz graph model \cite{wattsstrogatz}.
This is similar to previous work \cite{stier2019structural,xie2019exploring} in which random graph generators are used to construct a sparse neural network.
This can be seen as using a structural prior from social network theory.
Our interest with this approach is to determine whether the chosen prior contributes to the robustness of a network.
This has not only biological motivation but also motivation in analysing robustness measures in correlation to broader distributions of graph properties.

A single layer of a neural network is given as $z_l = \sigma(W_lz_{l-1}+B_l)$ for $l > 0$ and $z_0 = x$ being the input to the neural network, $W_l$ being the weights from layer $l-1$ to $l$, $B_l$ being a bias vector in layer $l$ and $\sigma$ being an activation function - we use Rectified Linear Unit \textit{ReLU} \footnote{An activation function that's commonly used in deep learning models, defined as $f(x)=\max(0,x)$.} throughout our experiments.
The prior structure as depicted in \autoref{fig:graph-induction} yields a binary connectivity mask $M_l$ in each layer based on the structures topological sorting and we can enforce a structure by reformulating $z_l$ as $\sigma(W_lM_lz_{l-1}+B_l)$.
Skip-layer connections in the structure introduce additional weight matrices from layer $s$ to $l$ and add for each combinations of a layer and preceding layers additive terms such that $z_l = \sigma(W_lM_lz_{l-1}+W_{s,l}M_{s,l}z_s+B_l)$.

On its left \autoref{fig:graph-induction} shows a Watts-Strogatz graph and a possible acyclic orientation of it in the middle.
The obtained directed acyclic graph on the right is used as a prior structure for a Sparse Neural Network.
An alternative way to obtain sparsity is through pruning.
In our case, the vertices determine neurons and we can naturally use the connectivity from one layer to the next as the mask $M_l$ or $M_{s,l}$ for $z_l$.
The first layer is fully connected to all vertices with no incoming edges and neurons in the last layer of which vertices have no outgoing edges in the graph are connected to the output dimension.
A layering of the graph can be recursively obtained through $ind^l(v): V\rightarrow\mathbb{N}$ with $v\mapsto max(\{ind^l(s) ~|~ (s,v)\in E(G)\})$.
Starting by assigning an index $l=0$ for all vertices with an in-degree equal to zero. Then, we iterate over the rest of the unsorted vertices: we only compute the index of vertex $v$ if and only if all the predecessors of $v$ have already been assigned an index. In that case, the index of $v$ is the maximum index of its predecessors incremented by $1$. 
The resulting sparse neural network can be denoted as $f$, being an image classifier on input $x$ with $f(x) = z_L(x)$ with $L$ being the last or maximal layer of the graph.
\vspace{-1em}


\section{Adversarial Robustness}\label{sec:adv-robustness}
Generating adversarial examples can be formalized as an optimization problem.
Let $f$ be the target image classifier, $x=(x_1,\cdots,x_n)$ be the input image where $x_i$ represent one pixel $p_i$, $f_l(x)$ being the probability of $x$ being classified with the correct label $l$, and $e(x)=(e_1,\cdots,e_n)$ being the additive adversarial perturbation vector.
The magnitude of the perturbation vector $e(x)$ is constrained by a limit $L$. Non-targeted attacks can be then formalized as in \autoref{eq:non-targeted-attack}.\vspace{-1em}

\begin{minipage}[b]{0.28\textwidth}
\begin{equation}\label{eq:non-targeted-attack}
	\begin{aligned}
		\min_{e(x)} \quad & f_{l} (x + e(x)) \\
		s.t. \quad & \lVert e(x) \rVert \leq L
	\end{aligned}
\end{equation}
\end{minipage}
~
\begin{minipage}[b]{0.34\textwidth}
\begin{equation}\label{eq:fgsm}
	\eta = \epsilon sign(\nabla_x C(\theta,x,y)) = e(x)
\end{equation}
\end{minipage}
~
\begin{minipage}[b]{0.28\textwidth}
\begin{equation}\label{eq:onepixel-fitness}
	\max_{c} \quad 1 - f_{l}(x(c))
\end{equation}
\end{minipage}

\begin{wrapfigure}{l}{0.3\textwidth}
    \centering
    \vspace{-3em}
    \includegraphics[width=0.3\textwidth]{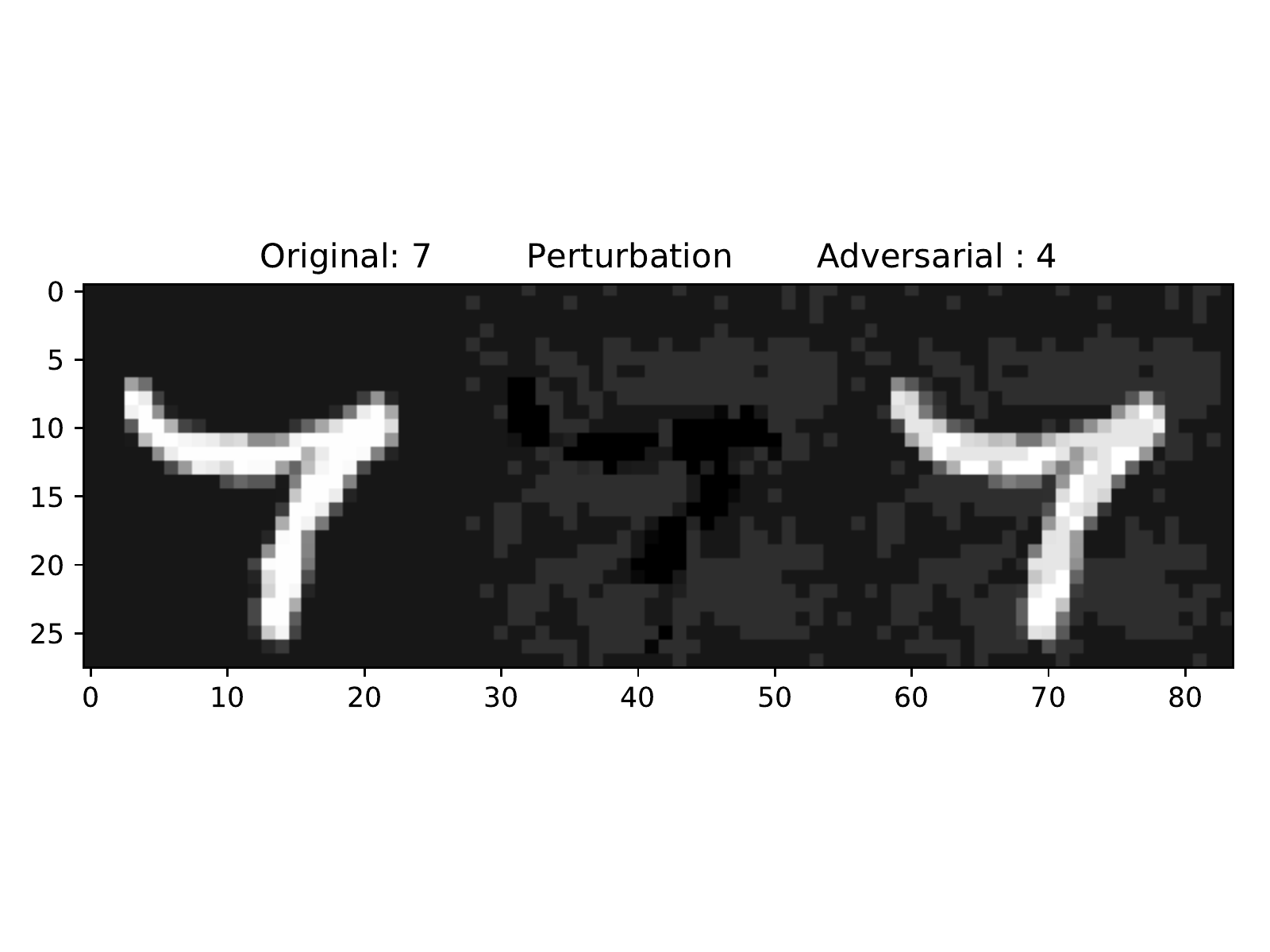}
    \vspace{-4em}
    \caption{An example of an image labeled as 7 perturbed with FGSM with $\epsilon = 0.1$ that was misclassified as 4 by a fully connected feed-forward network with three hidden layers (100, 50, 20).}
    \label{fig:fgsm_example}
    \vspace{-2em}
\end{wrapfigure}

\subparagraph{a. Fast Gradient Sign Method (FGSM)} was proposed by Goodfellow et al. in their paper \cite{goodfellow2014explaining}. 
The method was designed to generate adversarial examples by exploiting access to the numeric evalutation of the cost function derivative.
Let $x$ be the input to the model, $y$ the target label of $x$ and $C(\theta,x,y)$ be the cost function used to train the neural network. 
The adversarial perturbation $e(x)$, noted as $\eta$, is computed as in \autoref{eq:fgsm} in which $\nabla_x C()$ is the gradient of the cost function around the current value of the model parameters $\theta$ with respect to $x$ and $\epsilon$ is a small scalar value.
The adversarial example is the sum of the original sample and the perturbation, $\tilde{x} = x + \eta$.

\subparagraph{b. One Pixel} Su et al. \cite{su2017one} proposed a black-box attack by perturbing pixels, only one in our case, of the input images.
The only information required is the prior labels probabilities or simply the values of the output layer.
To compute adversarial examples, they use the concept of Differential Evolution (DE) \cite{storn1997differential} which is an evolution-based optimization algorithm.

Let $x$ be the target image to attack with class label $l$.
The algorithm for the non-targeted attack simply follows these steps:
\begin{enumerate}
	\item First, randomly generate an initial population of candidate solutions (perturbations) to feed into the differential evolution.
		The size of the initial population is a hyper-parameter $pop\_size$.
	\item Each candidate $c$ solution is encoded as a 3D vector $(p_x,p_y,I)$ with $p_{x,y}$ and $I$ being the coordinates and the intensity value of a random pixel in $Img$ respectively. 
		$I$ is to replace the original intensity value of the pixel.
	\item Iteratively, a new generation of candidates (or children) competes with the previous generation based on the following fitness function given in \autoref{eq:onepixel-fitness}.
\end{enumerate}

The maximum number of iteration is set by a hyper-parameter $max\_iter$.
The last surviving candidate is then used for the attack.
An early-stop criteria is activated when the true class is misclassified in the case of a non-targeted attack.\\
For the initial population of candidates, a Uniform distribution $\mathcal{U}(1,28)$ was used for the random $p_{x,y}$ coordinates (MNIST images have a size of 28x28) and a Gaussian distribution $\mathcal{N}(mean=128,std=127)$ was used for Intensity values.

The robustness of neural networks refers to the ability of these models to correctly classify adversarial examples.
Let $\mathcal{D}$ be the distribution of the test data samples, $f$ the image classifier, $x$, $l$, and $\tilde{x}$ an image, its associated label, and the perturbed image respectively.
To quantify the robustness of neural networks, we used three measures, namely the error rate, the average confidence and the average $\epsilon$.

\subparagraph{Error Rate} is the empirical error on the test data as shown in \autoref{eq:error-rate} or -- in other words -- the rate of successful misclassifications of the perturbed data samples.
It represent the probability of misclassification over the perturbed samples generated from the test data distribution.

\begin{minipage}[b]{0.28\textwidth}
\begin{equation}\label{eq:error-rate}
	\mathbb{P}_{x\sim\mathcal{D}}(f(\tilde{x})\neq l \,|\, f(x)=l)
\end{equation}
\end{minipage}
~
\begin{minipage}[b]{0.34\textwidth}
\begin{equation}\label{eq:confidence}
	\frac{1}{n} \sum_{x \in \mathcal{D}} f_{l^{''}}(\tilde{x})
\end{equation}
\end{minipage}
~
\begin{minipage}[b]{0.28\textwidth}
\begin{equation}\label{eq:average-epsilon}
	\bar{\epsilon} = \mathbb{E}(\epsilon_x) = \frac{1}{n} \sum_{x \in \mathcal{D}} \epsilon_x
\end{equation}
\end{minipage}

\subparagraph{Average confidence} represents the expected confidence -- the probability of the predicted class -- of the classifier over misclassifications of adversarial examples.
With $n$ being the number of successfully misclassified adversarial images and $l^{''}$ the wrongly assigned label for $\tilde{x}$ the confidence is given in \autoref{eq:confidence}.

\subparagraph{Average $\epsilon$ (FGSM)} is used only for the FGSM attack, in which $\epsilon$ is not fixed during the attacks.
For every image $x$, $\epsilon_x$ starts from a small value of $0.001$ and keeps increasing by $0.01$ until the perturbed image gets misclassified. 
This represent a point-wise robustness measure where we measure how resistant the model is to the scaled additive perturbation $\epsilon_x sign(\nabla_x C(\theta,x,y))$ on the datapoint $x$ by increasing $\epsilon_x$ until the model misclassifies the perturbed $x$. 
The expectation of $\epsilon$ over all successful misclassifications is given in \autoref{eq:average-epsilon}.


\section{Experimental Design}\label{sec:experimental_design}
We conducted multiple experiments in which we correlate structural properties of the Sparse Neural Network (SNN) with robustness measures.
The first experiment entails generated Watts-Strogatz graphs induced as a structural prior to Sparse Neural Networks.
In an ablation study on the weight initialization we excluded at least the possibility of that hyperparameter as a significant influence on different robustness measures.

\paragraph{Training}
We use Adam as optimizer with learning rate $\eta = 1e^{-3}$, $\beta_1 = 0.9$, $\beta_2 = 0.999$, $\epsilon = 1e^{-8}$, $\sigma = ReLU$ as activation function and $30$ training epochs.
We trained on the image classification problem MNIST \cite{lecun2010mnist} with $d_{in} = 28\times 28 = 784$.

\paragraph{Weights Initialization}
To show that not every arbitrary hyperparameter has an immediate effect on robustness, we used six different \textit{weight initializations}:
The Glorot/Xavier \cite{glorot2010understanding} initialization with using both a normal \textbf{(G\_N)} and a uniform distribution \textbf{(G\_U)}.
For both, we used $gain=\sqrt{2}$ as parameter.
The Kaiming/He \cite{he2015delving} initialization with using both a normal \textbf{(He\_N)} and uniform distribution \textbf{(He\_U)}.
Again, with parameters $a=0$, $mode=fan_{in}$ and $gain=\sqrt{2}$.
The normal distribution \textbf{(N)} initialization $\mathcal{N}(mean,std)$ with $mean = 0$ and $std = 0.1$.
And the uniform distribution \textbf{(U)} initialization $\mathcal{U}(a,b)$ with lower bound $a=-0.1$ and upper bound $b=0.1$.

\paragraph{Adversarial Attacks}
As evaluation metrics we used the two \textit{adversarial attacks} presented in \autoref{sec:adv-robustness}:
(1) FGSM with $\epsilon = 0.1$ as a parameter. 
This attack was applied to the whole test set of $10,000$ samples.
(2) One Pixel with $max_{iter} = 500$ and $pop_{size} = 500$ for the differential evolution. 
We only applied non targeted attack mode to \textbf{a subset of $100$} test images due to high demand on computation resources for this attack.

\paragraph{Correlation Analysis}
Our correlation analysis in our study consists of two rank correlation coefficients for each pair of the variables in question, \textbf{Spearman}-$\rho$ \cite{spearman1904proof} and \textbf{Kendall}-$\tau$ \cite{kendall1948rank}.
These non-parametric tests provide measures of the degree of association (Spearman) and the strength of dependence (Kendall) between two variables.
Both correlation statistics do not require any assumptions about the distribution of the data and are not limited to linear relationships. 
Cohen's standard \cite{cohen2014applied} is used to evaluate the correlation coefficient to determine the strength of the relationship in which .10 - .29 denote weak or small, .30 - .49 medium or moderate and .50 and above a large association.

\subparagraph{Graphs Dataset}
To study the robustness of SNNs, we generated a set of $100$ random graphs $\{g \rightarrow WS(\cdots)\}$ using the Watts-Strogatz model, induced each graph as a structural prior of the network $N(g)$, and trained the resulting models repeatedly for six iterations with six different initialization of their weigths $W$. 
We then evaluated the resulting 600 models to study the existence of any correlation between their robustness measures and their graph properties.

The generation of the $100$ graphs was done through a limited \textbf{grid search} over the generator parameters:
number of nodes or the size of the lattice $size$, distance within which two vertices will be connected $nei$ and the probability of rewiring an edge $p$.
The condition on which a graph $g$ gets added to the set was that the resulting network has a number of parameters varying from $50k$ to $91k$.
The reason behind this range of values was to make the experiment complimentary to the pruning evaluation in terms of number of parameters between the models tested.
We chose $size \in \{250,300,350,400,500\}$, $nei \in \{2,4,6,8,10,20\}$ and $p \in \{0.5,0.6,0.7,0.8,0.9\}$.
An important note to recall here is that the $size$ mentioned above only represent the number of nodes of the hidden layers of the output network.

\subparagraph{Graph Properties Distribution}
Having this set of SNNs, we investigated the distributions of their graph properties to extract any irregular statistical characteristics of the set of models we are using for the robustness evaluation.

\begin{figure*}[h!]
	\vspace{-1.5em}
	\centering
	\subfloat[
		Distribution of the number of parameters of the generated models. A model is selected during the generation process when its number of parameters falls within the range of $50,000$-$91,000$.
		\label{fig:snnsparametersdist}
	]{\includegraphics[width=0.4\textwidth,valign=m]{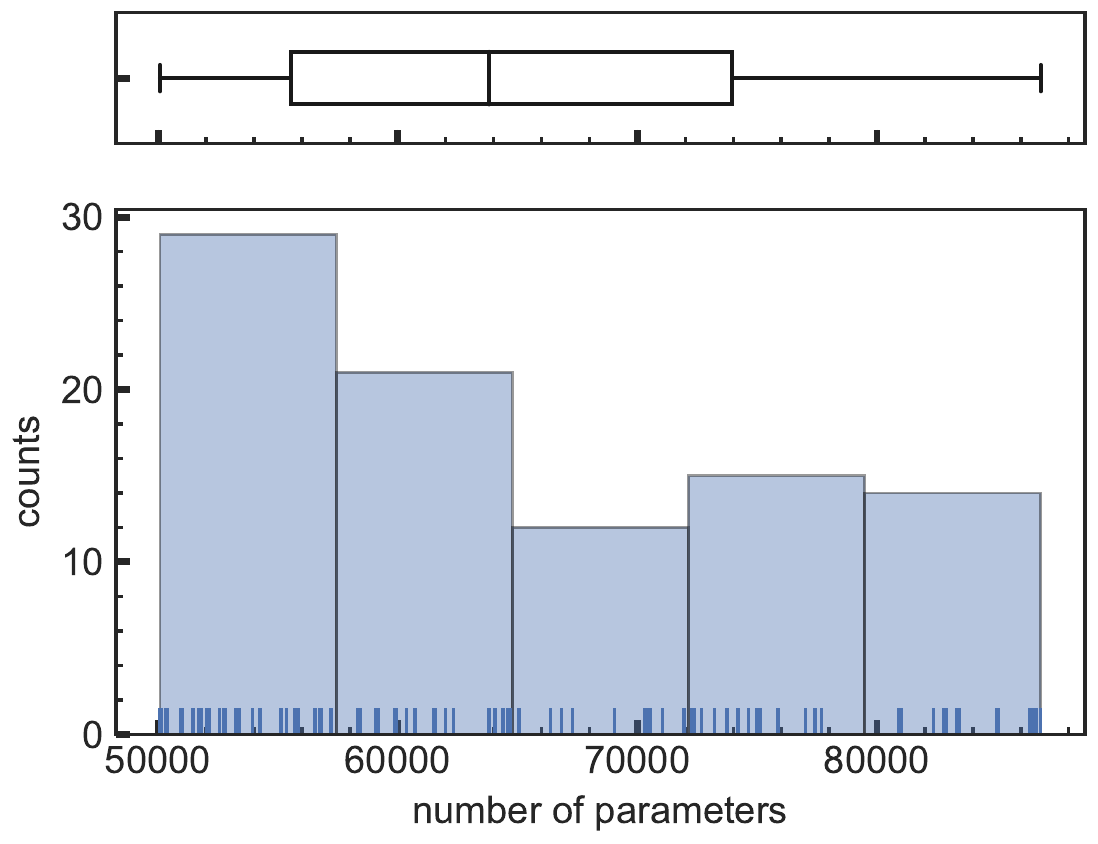}}
	\quad
	\subfloat[
		Distribution of the number of parameters with respect to the Watts-Strogatz model hyper-parameters $size$, $p$ and $nei$ in sub-figure (1), (2), and (3-4) respectively. Across the 4 sub-figures, blue colour refers to models generated using a neighborhood $nei=2$ while orange refers to $nei=4$.
		\label{fig:generatorparamsinv}
	]{\includegraphics[width=0.45\textwidth,valign=m]{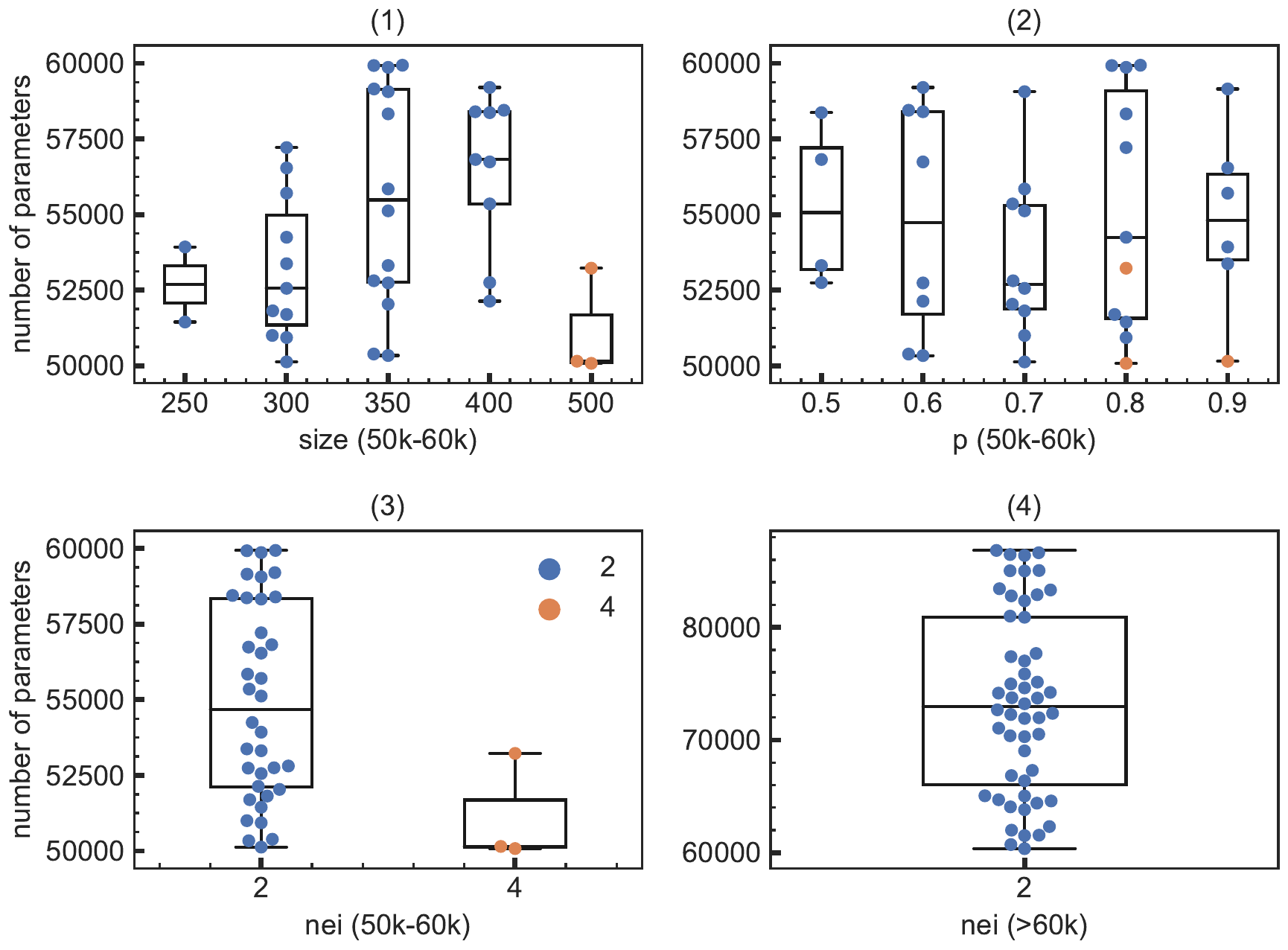}}
	\caption{Distributions of properties of the resulting models from graphs generated by uniformly sampled hyper-parameters for the graph generator.}
	\vspace{-1em}
\end{figure*}
Figure \ref{fig:snnsparametersdist} shows that the distribution of the SNNs with respect to their number of parameters is slightly right-skewed and almost evenly spread over the range of $60,000 - 91,000$. 
However, within the first $10k$ of the spectrum, we observe that there are more samples -- approximately $30\%$ of all the networks -- than the rest of the bins.

To answer the question whether this is related to the implemented grid search method, we looked at the distribution of this group of models with respect to the generator's parameters (see (1) and (2) in \autoref{fig:generatorparamsinv}).
It shows that it is not the case for both $size$ and rewiring probability $p$ parameters since they have a spreaded distribution.
This confines the issue to the $nei$ parameter.
However, when we compare the distribution of this group of graphs to the distribution of the rest of the set with respect to $nei$, as in (2) and (3) \autoref{fig:generatorparamsinv}, we can fairly say that both groups have the same parameter value ($nei=2$) overall which answers the aforementioned question.

This small discrepancy in the distribution of the number of parameters is then coming from the randomness of the graph generator solely and should not affect the evaluation process.

Aside from the distribution of the samples' number of parameters, we wanted to look at the distribution of graph properties among the set of graphs.

\begin{figure}[tb]
	\centering
	\subfloat[
		The density is the proportion between the number of edges and all possible edges.
		\label{fig:distributions-density}
	]{\includegraphics[width=0.24\textwidth]{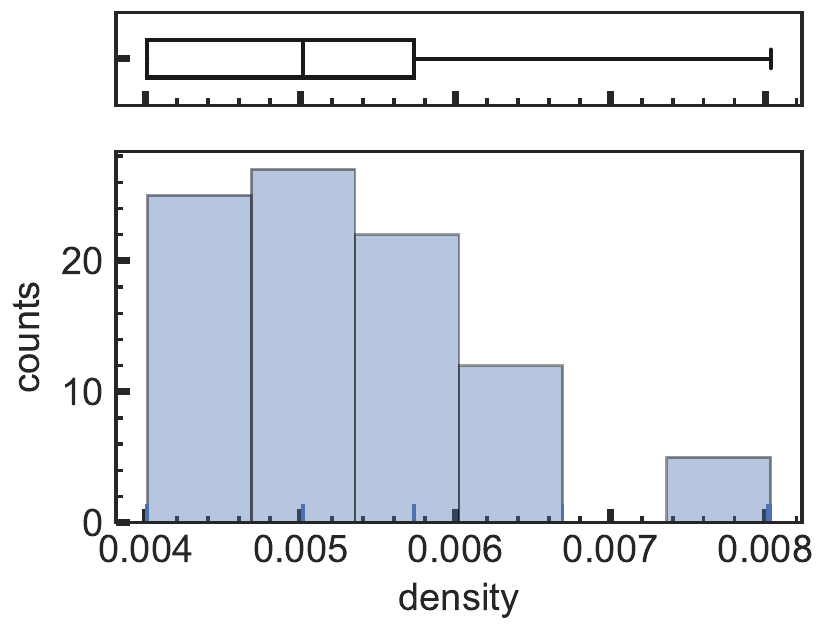}}
	~
	\subfloat[
		The average eccentricity is the mean over all longest paths $\epsilon(v)$ between $v$ and any other vertex in the graph.
		\label{fig:distributions-avgeccentricity}
	]{\includegraphics[width=0.24\textwidth]{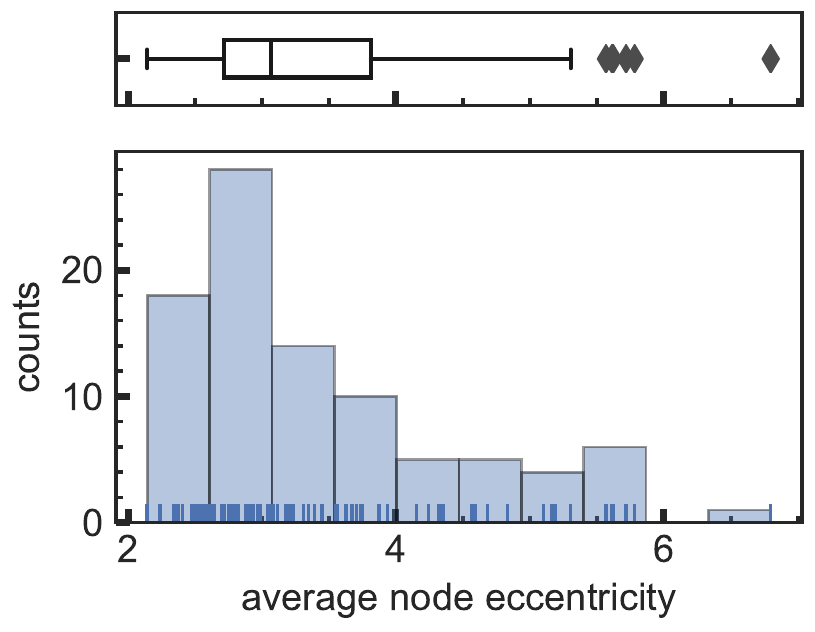}}
	~
	\subfloat[
		The average path length is the mean over all distances of pairs of vertices $s, t\in V$ in a graph.
		\label{fig:distributions-pathlength}
	]{\includegraphics[width=0.24\textwidth]{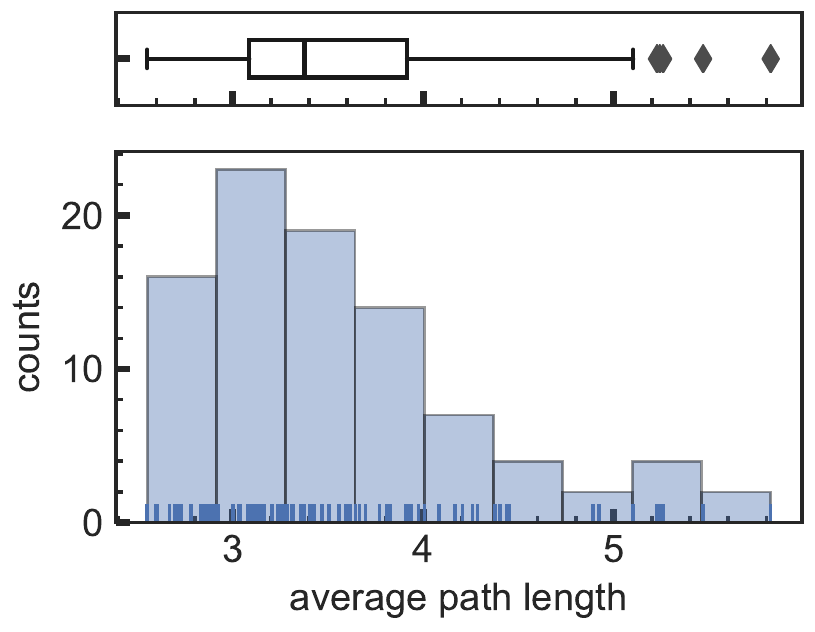}}
	~
	\subfloat[
		Average betweenness is the mean of the betweenness centrality.
		\label{fig:distributions-avgbetweenness}
	]{\includegraphics[width=0.24\textwidth]{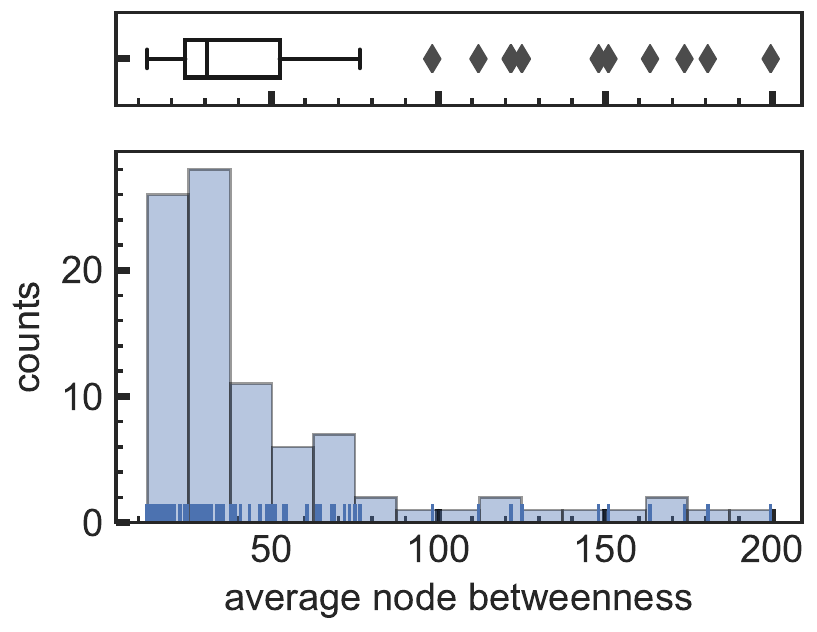}}
	\caption{
		Distributions of structural properties across all graph samples.
		It is noteworthy, that we enforced a close-to-uniform distribution on the number of parameters as it can be seen in \autoref{fig:snnsparametersdist} (which is highly correlated with the vertices of the underlying graph) but at the same time other parameters in the sampled space are not following such a simple distribution.
	}
	\label{fig:distributions}
	\vspace{-2em}
\end{figure}

\autoref{fig:distributions} shows distributions of properties on which we found interesting correlations later on.
All of the distributions are uni-modal and right-skewed. 
For the case of the diameter, the average path length and the average node eccentricity, the spread of values is wider than the other properties.
More particularly, the distribution of the average node betweeness as seen in \autoref{fig:distributions-avgbetweenness}) and closeness seem to be condensed in a small chunk of the range spread -- i.e. $62\%$ of the samples lies within the first quarter of the values' range --, insinuating that these two properties will not provide a constructive insight on the models' behavior.

\section{Evaluation and Results}\label{sec:results}
We provide information on performance and robustness measures of the Sparse Neural Networks and a correlation analysis between these measures and graph theoretic properties.

\begin{figure*}[hbt]
	\vspace{-1.5em}
	\centering
	\subfloat[
		The distribution of the robustness (or fail rate) of all the models with respect to the six initialization methods against the FGSM and the One Pixel attacks.
		\label{fig:distri_robust_init}
	]{\includegraphics[width=0.4\textwidth,valign=m]{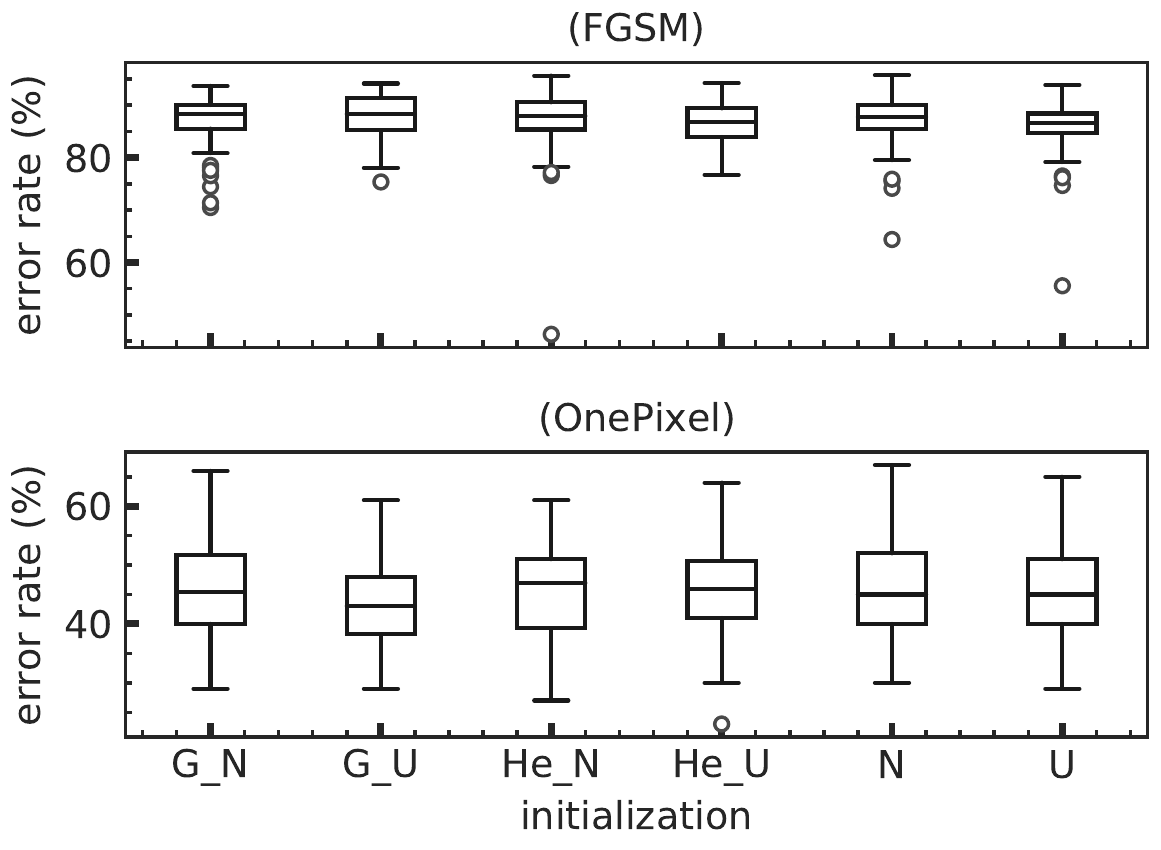}}
	\qquad
	\centering
	\subfloat[
		The distribution of f1-measure on MNIST with respect to the six initialization methods.
		\label{fig:distacc}
	]{\includegraphics[width=0.4\textwidth,valign=m]{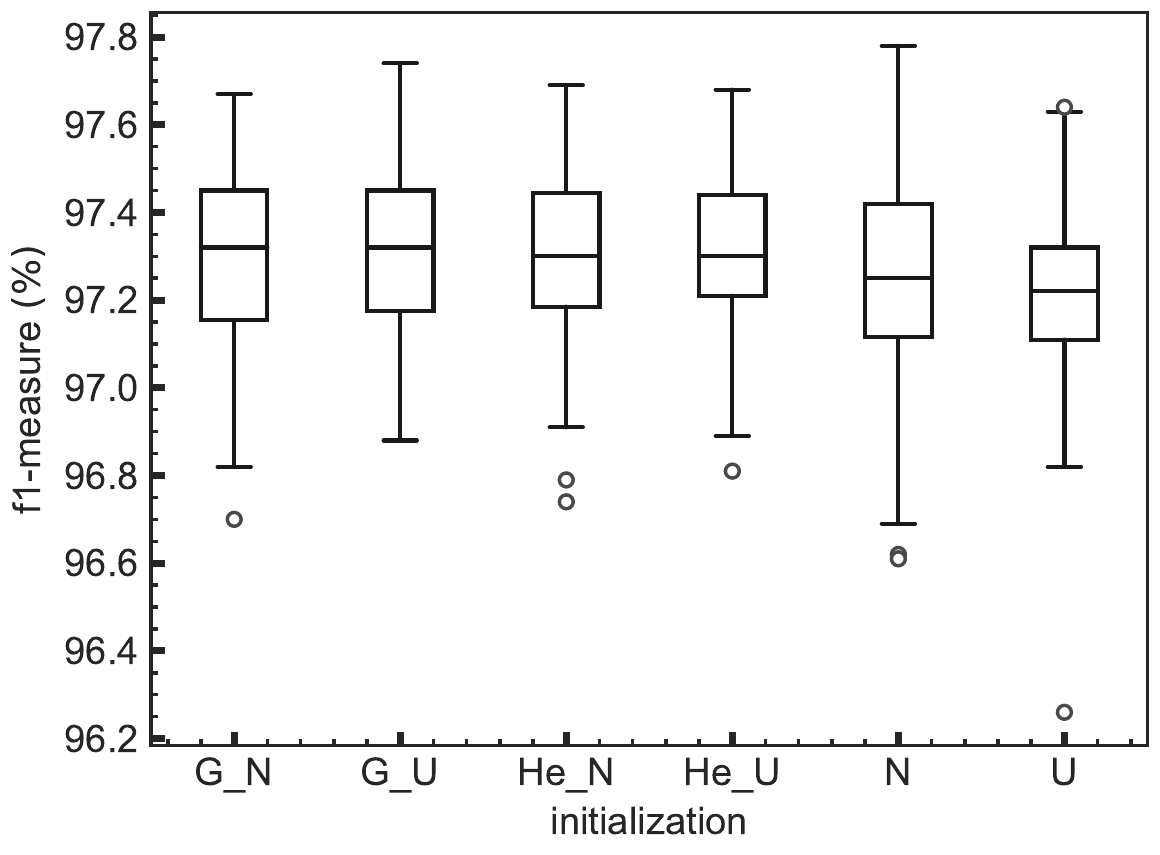}}
	\caption{Distributions of performance and robustness measures of all models with respect to weights' initialization methods.}
	\vspace{-1.5em}
\end{figure*}

\paragraph{Performance and Robustness Measures}
In \autoref{fig:distri_robust_init}, distributions of robustness measures for both FGSM and One Pixel attack with respect to the six different initialization methods is shown.
It depicts the general effect of the weights and biases' initial values on the robustness measures.
Clearly, the weight initialization in our ablation study has no influence on final robustness measures.
As it can be seen in \autoref{fig:distacc} the initialization methods also exhibit stable distributions for the f1-measure which is overall between 96.6 and 97.8 with a mean of 97.3 across our dataset.
F1-measures for our pruned SNNs range between 97.5 and 97.9 with a observed mean of about 97.7.
We see the reasoning behind this higher mean in the re-training phases after pruning steps.

\begin{figure*}[tb]
	\vspace{-1.5em}
	\centering
	\subfloat[
		The distribution of the average perturbation scale factor $\bar{\epsilon}$ on misclassfication of adversarial examples of FGSM with respect to the six different initialization.
		\label{fig:disteps}
	]{\includegraphics[width=0.4\textwidth,valign=m]{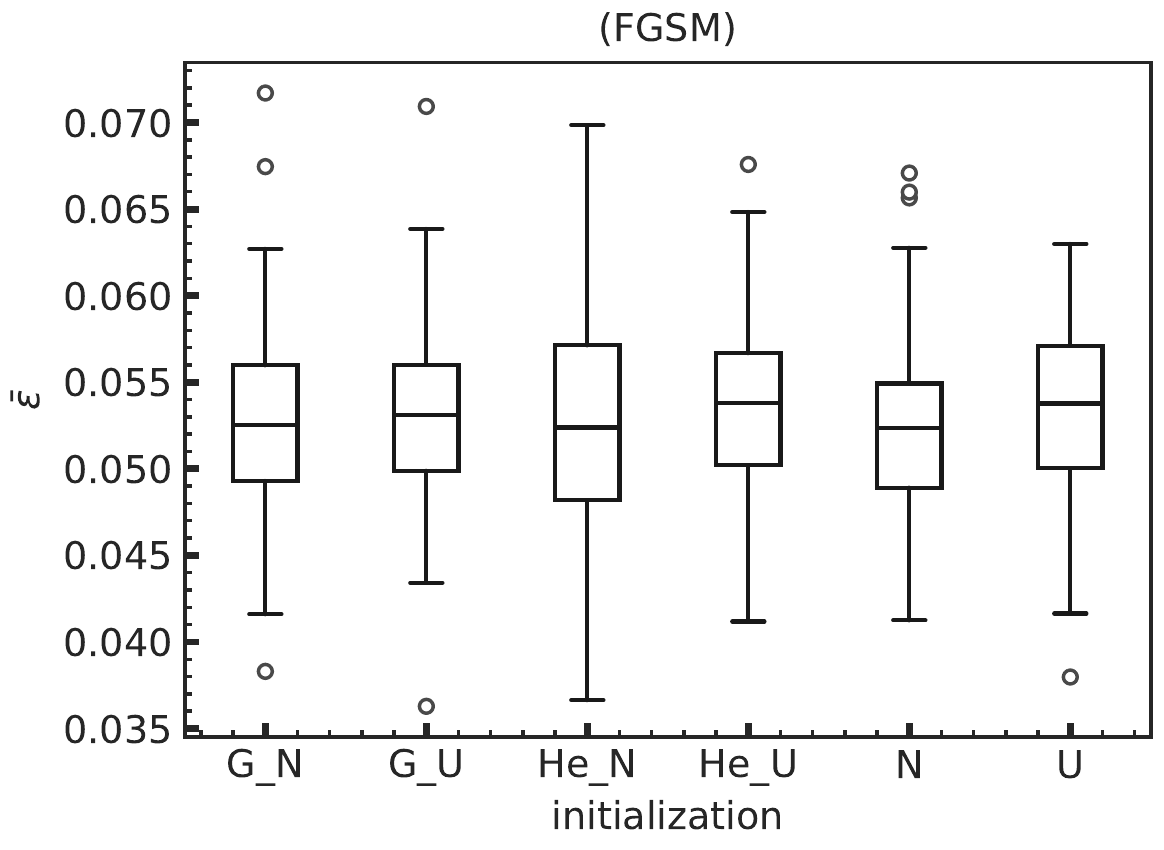}}
	\qquad
	\subfloat[
		The distribution of the average model confidence on misclassfication of adversarial examples, FGSM and One Pixel, with respect to the six different initialization.
		\label{fig:distavgconf}
	]{\includegraphics[width=0.4\textwidth,valign=m]{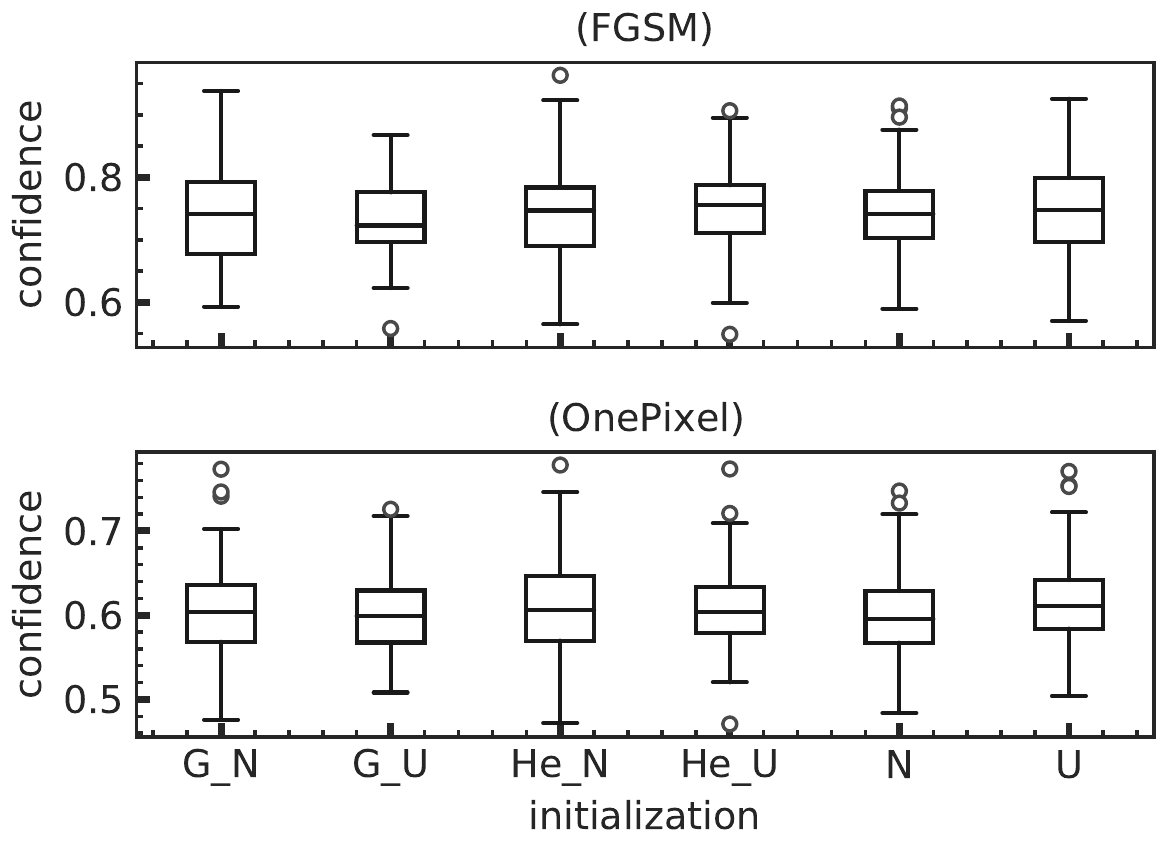}}
	\caption{Distributions of robusntess measures, confidence values and average epsilons, of all models with respect to initialization methods.}
	\vspace{-1em}
	
\end{figure*}

We observe that for 100 samples a large standard deviation in robustness comes from normally initialized networks such as He-normal and normal-initialization, whereas more stable robustness distributions with low $std$ come from uniform initializations such as Xavier and Kaiming uniform.
We also notice that the interquartile range ($\approx 5\%$) across all initialization is so low compared to the spread of values.
This could be due to the discrepancy we discussed in the previous section on graph properties.
For One Pixel, the distributions seem to be similar in terms of the spread of robustness values.

The distributions of the average epsilon and the average confidence are respectively in figures\autoref{fig:disteps} and\autoref{fig:distavgconf}. 

\clearpage
\begin{wraptable}{r}{0.5\textwidth}

\caption{Correlation coefficients between robustness evaluations and the three graph properties with the highest $\rho$ and $\tau$.}\label{tab:corranalysis}
    \centering
    \resizebox{0.5\textwidth}{!}{
        \begin{tabular}{l l l l l l}
        & \multicolumn{3}{c }{FGSM} & \multicolumn{2}{c }{One Pixel} \\
        \cline{2-6} & error rate & confidence & $\bar{\epsilon}$ & error rate & confidence\\
        \midrule
        Number of parameters & \makecell{$\rho=\textbf{-0.16}$\\$\tau=\textbf{-0.11}$} & \makecell{$\rho=0.38$\\$\tau=0.25$} & \makecell{$\rho=\textbf{0.43}$\\$\tau=\textbf{0.30}$} & \makecell{$\rho=\textbf{-0.58}$\\$\tau=\textbf{-0.40}$} & \makecell{$\rho=\textbf{0.40}$\\$\tau=\textbf{0.27}$} \\
        \midrule
        Density & \makecell{$\rho=\textbf{0.32}$\\$\tau=\textbf{0.24}$} & \makecell{$\rho=\textbf{-0.66}$\\$\tau=\textbf{-0.52}$} & \makecell{$\rho=-0.17$\\$\tau=-0.13$} & \makecell{$\rho=0.29$\\$\tau=0.21$} & \makecell{$\rho=\textbf{-0.63}$\\$\tau=\textbf{-0.49}$} \\
        \midrule
        Average path length & \makecell{$\rho=-0.15$\\$\tau=-0.11$} & \makecell{$\rho=\textbf{0.37}$\\$\tau=\textbf{0.25}$} & \makecell{$\rho=\textbf{-0.30 }$\\$\tau=\textbf{-0.22}$} & & \\
        \midrule
		Average eccentricity & & & & \makecell{$\rho=\textbf{0.33}$\\$\tau=\textbf{0.21}$} & \makecell{$\rho=0.21$\\$\tau=0.14$} \\
        \bottomrule
        \end{tabular}
        }
\vspace{-1em}
\end{wraptable}

\paragraph{Correlation of Robustness and Graph Properties}
We conducted an empirical correlation analysis between every pair of robustness measures and graph properties.

Due to the skewness existing in the distributions of the graph properties and the high variance of the robustness measures of our models, a preprocessing of the results is needed before studying any potential relationships between the measured values. 
The first step is to detect and discard any outlier in the distribution of the robustness measures. Any model with robustness measures that fall below $Q1 - 1.5IQR$ or above $Q3 + 1.5IQR$ of the distributions is to be discarded.  
After that, we computed the expectation or the mean of the $6$ collected values of the robustness measures for every sample network.

\begin{wraptable}{l}{0.5\textwidth}
\caption{Correlation coefficients between robustness evaluations and the chosen graph properties in \autoref{tab:corranalysis} of the fully connected model over 20 pruning steps.}\label{tab:corranalysispruning}

    \centering
    \resizebox{0.5\textwidth}{!}{
    \begin{tabular}{l l l l l l}
        & \multicolumn{3}{c }{FGSM} & \multicolumn{2}{c }{One Pixel} \\
        \cline{2-6} & error rate & confidence & $\bar{\epsilon}$ & error rate & confidence\\
        \midrule
        Number of parameters & \makecell{$\rho=-0.51$\\$\tau=-0.35$} & \makecell{$\rho=-0.64$\\$\tau=-0.45$} & \makecell{$\rho=0.66$\\$\tau=0.46$} & \makecell{$\rho=-0.62$\\$\tau=-0.44$} & \makecell{$\rho=0.36$\\$\tau=-0.39$} \\
        \midrule
        Density & \makecell{$\rho=-0.51$\\$\tau=-0.35$} & \makecell{$\rho=-0.64$\\$\tau=-0.45$} & \makecell{$\rho=0.66$\\$\tau=-0.46$} & \makecell{$\rho=-0.62$\\$\tau=-0.44$} & \makecell{$\rho=-0.56$\\$\tau=-0.39$} \\
        \midrule
        Average path length & \makecell{$\rho=0.51$\\$\tau=0.35$} & \makecell{$\rho=0.64$\\$\tau=0.45$} & \makecell{$\rho=-0.66$\\$\tau=-0.46$} & \makecell{$\rho=0.62$\\$\tau=0.44$} & \makecell{$\rho=0.56$\\$\tau=0.39$}\\
        \midrule
	Average eccentricity & & & & \makecell{$\rho=\textbf{-0.34}$\\$\tau=\textbf{-0.22}$} & \makecell{$\rho=0.26$\\$\tau=0.17$} \\
        \bottomrule
    \end{tabular}
    }
\end{wraptable}

The results of the correlation analysis are presented in \autoref{tab:corranalysis} for the two adversarial attacks. 
We then only considered the graph properties with the two largest coefficients for each robustness metric.
We observe that for the error rate, the density of the network and the number of its parameters have the highest correlation coefficients $\rho$ and $\tau$. 
Both properties seem to have a weak relationship with the error rate but in opposite directions since the error rate has a negative weak association with the density ($\rho=-0.26 < 0$) and a very weak positive one ($\rho=0.16>0$) with the number of parameters.

\begin{figure}[tb]
	\vspace{-1.5em}
	\centering
	\subfloat[
		Error rate in relation to density of the underlying graph.
		We observe a weak negative correlation of these properties.
		\label{fig:robus_density}
	]{\includegraphics[width=0.35\textwidth]{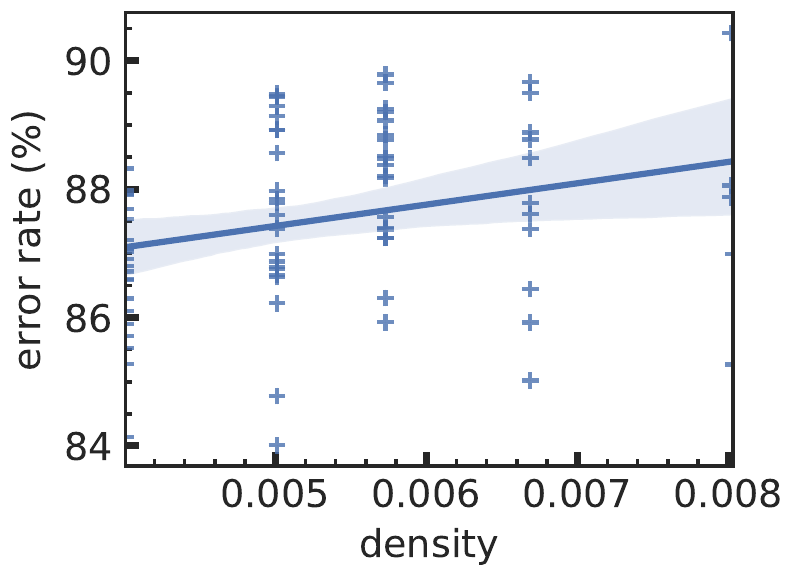}}
	~
	\subfloat[
		Average confidence in relation to the density of the underlying graph.
		A large negative correlation can be observed.
		\label{fig:confdensity}
	]{\includegraphics[width=0.35\textwidth]{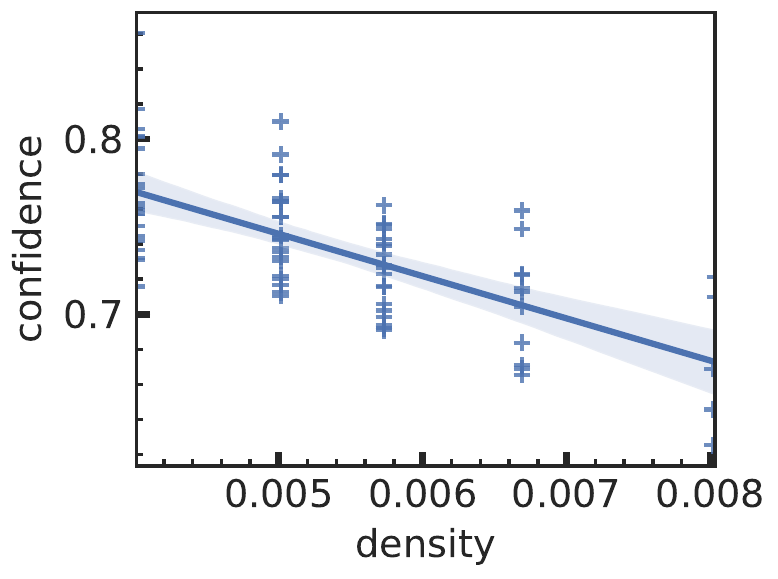}}
	\caption{The relational scatter plots between density and both the error rate and the average confidence for One Pixel.}
	\vspace{-1em}
\end{figure}

Figure \ref{fig:robus_density} shows the error rate with respect to the density in which we can observe that the weak association is almost non significant.
However, the average confidence measure seems to have a large or strong monotonically decreasing relationship with the density (a correlation coefficient $\rho=-0.64$) and a moderate increasing association with the average path length of the models ($\rho=0.39$).

\begin{figure}[bt]
	\vspace{-1.5em}
	\centering
	\subfloat[
		Average $\epsilon$ related to the number of parameters of the models.
		\label{fig:eps_params}
	]{\includegraphics[width=0.35\textwidth]{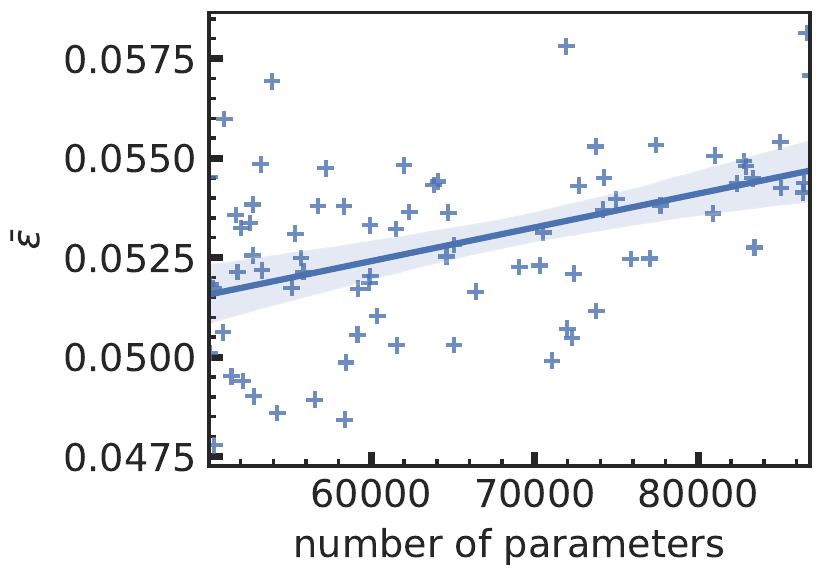}}
	\quad
	\subfloat[
		The error rate for one pixel attack and the number of parameters of the models.
		\label{fig:robust_params_op}
	]{\includegraphics[width=0.35\textwidth]{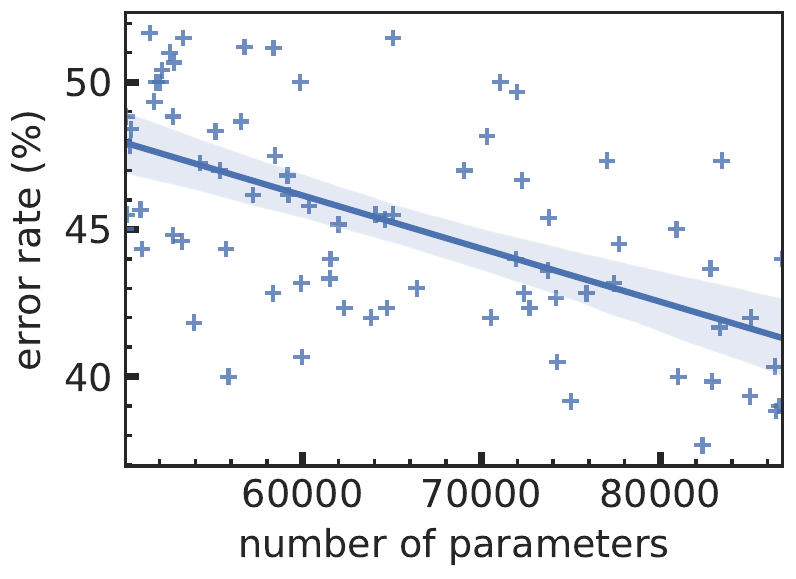}}
	\caption{The relational scatter plots between number of parameters and both the error rate and the average noise scale factor for FGSM.}
        \vspace{-1.5em}
\end{figure}

The relation between the average confidence and the density (see \autoref{fig:confdensity}) revokes the relation between the error rate measure and the density. 
The two measures should be inversely proportional in case they have a linear relationship.
Average $\epsilon$ has two moderate relationships, a positive one with the number of edges (with $\rho=0.42$) and a negative one with average path length ($\rho=-0.34$).
Inspecting this relationship in \autoref{fig:eps_params} confirms our observation.

The error rate for one pixel attack shows a similar and stronger positive monotonicity with the number of parameters which can be seen in the correlation coefficient ($\rho=0.55$) and in the relational plot (see \autoref{fig:robust_params_op}). 

\autoref{tab:corranalysispruning} includes the correlation values we found when evaluating the robustness a fully connected model after each pruning step. 
The goal was to have a point of reference to compare with the previous results. 
Random pruning is simply a method that prunes a proportion $\alpha$ of the weights randomly after each pruning step.  
The hidden layers of the model are composed by $50$, $100$, $100$ and $50$ nodes respectively. 
We only applied the pruning on connections between the hidden neurons in order to make it comparable with SNNs experiment.
Additionally, graph properties were only computed for the graph obtained from the hidden structure of the network.  
We notice that most of the coefficients have similar intensity which we can be due to a dependency introduced by the pruning step or simply due to lack of data. 
An interesting observation is the fact that random pruning seems to cause a negative effect on the robustness of the model. 

\section{Conclusion}\label{sec:conclusion}
We presented a method for inducing graphs as structural priors of Sparse Neural Networks (SNNs).
SNNs were initialized with six different distributions, trained and evaluated on MNIST with f1-score as performance measure and error rate, confidence and average-$\epsilon$ as robustness measures against two adversarial attacks, FGSM and One Pixel attack.

Surprisingly, the \textit{density} shows both a negative correlation with the robustness and the average confidence of SNNs. 
If hypothetically there would be a clear linear relationship between the two robustness measures, we would expect them to have an opposite correlations with the density.
Our interpretation then is that there is a non-linear relationship between both notions of robustness.
Moreover, we observe that lower densities lead to more robust SNNs. 
The \textit{average path length} shows a positive correlation with the average confidence score. 
Additionally, the \textit{average eccentricity} shows a negative correlation with the robustness measure.
Both properties encode the connectivity between neurons and we observe that lower connectivity correlates with higher robustness measures.
We also found there is clearly a negative effect on the robustness after each pruning step -- something which is not as obviously observed in experiments with complex initialized structures, promising an effect of other structural properties on the robustness.

However, even with the fact that we showed an existing relationship between the robustness of SNNs and their graph properties, some conflicts and inconsistencies in the results along with insufficient number of SNNs tested makes it inaccurate to draw strong conclusions from it which, however, motivates us to further investigate this complex relationship.
Further studies could lead to develop our hypothesis further for a new research path such as diversifying the used graph generator and drilling into the correlation analysis by investigating non-linear and multi-variate relations.


\bibliographystyle{unsrtnat}

\begin{thebibliography}{23}
\providecommand{\natexlab}[1]{#1}
\providecommand{\url}[1]{\texttt{#1}}
\expandafter\ifx\csname urlstyle\endcsname\relax
  \providecommand{\doi}[1]{doi: #1}\else
  \providecommand{\doi}{doi: \begingroup \urlstyle{rm}\Url}\fi

\bibitem[Akhtar and Mian(2018)]{akhtar2018threat}
Naveed Akhtar and Ajmal Mian.
\newblock Threat of adversarial attacks on deep learning in computer vision: A
  survey.
\newblock \emph{arXiv preprint arXiv:1801.00553}, 2018.

\bibitem[Thom(2015)]{thom2015sparse}
Markus Thom.
\newblock \emph{Sparse neural networks}.
\newblock PhD thesis, Universit{\"a}t Ulm, 2015.

\bibitem[LeCun et~al.(1995)LeCun, Bengio, et~al.]{lecun1995convolutional}
Yann LeCun, Yoshua Bengio, et~al.
\newblock Convolutional networks for images, speech, and time series.
\newblock \emph{The handbook of brain theory and neural networks},
  3361\penalty0 (10):\penalty0 1995, 1995.

\bibitem[Louizos et~al.(2017)Louizos, Welling, and Kingma]{louizos2017learning}
Christos Louizos, Max Welling, and Diederik~P Kingma.
\newblock Learning sparse neural networks through $ l\_0 $ regularization.
\newblock \emph{arXiv preprint arXiv:1712.01312}, 2017.

\bibitem[Frankle and Carbin(2018)]{frankle2018lottery}
Jonathan Frankle and Michael Carbin.
\newblock The lottery ticket hypothesis: Finding sparse, trainable neural
  networks.
\newblock \emph{arXiv preprint arXiv:1803.03635}, 2018.

\bibitem[Elsken et~al.(2018)Elsken, Metzen, and Hutter]{elsken2018neural}
Thomas Elsken, Jan~Hendrik Metzen, and Frank Hutter.
\newblock Neural architecture search: A survey.
\newblock \emph{arXiv preprint arXiv:1808.05377}, 2018.

\bibitem[Stier and Granitzer(2019)]{stier2019structural}
Julian Stier and Michael Granitzer.
\newblock Structural analysis of sparse neural networks.
\newblock \emph{Procedia Computer Science}, 159:\penalty0 107--116, 2019.

\bibitem[Xie et~al.(2019)Xie, Kirillov, Girshick, and He]{xie2019exploring}
Saining Xie, Alexander Kirillov, Ross Girshick, and Kaiming He.
\newblock Exploring randomly wired neural networks for image recognition.
\newblock In \emph{Proceedings of the IEEE International Conference on Computer
  Vision}, pages 1284--1293, 2019.

\bibitem[Goodfellow et~al.(2014)Goodfellow, Shlens, and
  Szegedy]{goodfellow2014explaining}
Ian~J Goodfellow, Jonathon Shlens, and Christian Szegedy.
\newblock Explaining and harnessing adversarial examples.
\newblock \emph{arXiv preprint arXiv:1412.6572}, 2014.

\bibitem[Fawzi et~al.(2018)Fawzi, Fawzi, and Frossard]{fawzi2018analysis}
Alhussein Fawzi, Omar Fawzi, and Pascal Frossard.
\newblock Analysis of classifiers’ robustness to adversarial perturbations.
\newblock \emph{Machine Learning}, 107\penalty0 (3):\penalty0 481--508, 2018.

\bibitem[Fawzi et~al.(2017)Fawzi, Dezfooli, and Frossard]{fawzi2017geometric}
Alhussein Fawzi, SM~Moosavi Dezfooli, and Pascal Frossard.
\newblock A geometric perspective on the robustness of deep networks.
\newblock \emph{Institute of Electrical and Electronics Engineers, Tech. Rep},
  2017.

\bibitem[Szegedy et~al.(2013)Szegedy, Zaremba, Sutskever, Bruna, Erhan,
  Goodfellow, and Fergus]{szegedy2013intriguing}
Christian Szegedy, Wojciech Zaremba, Ilya Sutskever, Joan Bruna, Dumitru Erhan,
  Ian Goodfellow, and Rob Fergus.
\newblock Intriguing properties of neural networks.
\newblock \emph{arXiv preprint arXiv:1312.6199}, 2013.

\bibitem[Kurakin et~al.(2016)Kurakin, Goodfellow, and
  Bengio]{kurakin2016adversarial}
Alexey Kurakin, Ian Goodfellow, and Samy Bengio.
\newblock Adversarial examples in the physical world.
\newblock \emph{arXiv preprint arXiv:1607.02533}, 2016.

\bibitem[Diestel(2017)]{diestel2017graph}
Reinhard Diestel.
\newblock \emph{Graph Theory}.
\newblock Springer-Verlag, 2017.

\bibitem[Watts and Strogatz(1998)]{wattsstrogatz}
Duncan~J Watts and Steven~H Strogatz.
\newblock Collective dynamics of ‘small-world’networks.
\newblock \emph{nature}, 393\penalty0 (6684):\penalty0 440, 1998.

\bibitem[Su et~al.(2017)Su, Vargas, and Kouichi]{su2017one}
Jiawei Su, Danilo~Vasconcellos Vargas, and Sakurai Kouichi.
\newblock One pixel attack for fooling deep neural networks.
\newblock \emph{arXiv preprint arXiv:1710.08864}, 2017.

\bibitem[Storn and Price(1997)]{storn1997differential}
Rainer Storn and Kenneth Price.
\newblock Differential evolution--a simple and efficient heuristic for global
  optimization over continuous spaces.
\newblock \emph{Journal of global optimization}, 11\penalty0 (4):\penalty0
  341--359, 1997.

\bibitem[LeCun et~al.(2010)LeCun, Cortes, and Burges]{lecun2010mnist}
Yann LeCun, Corinna Cortes, and CJ~Burges.
\newblock Mnist handwritten digit database.
\newblock \emph{AT\&T Labs [Online]. Available: http://yann. lecun.
  com/exdb/mnist}, 2, 2010.

\bibitem[Glorot and Bengio(2010)]{glorot2010understanding}
Xavier Glorot and Yoshua Bengio.
\newblock Understanding the difficulty of training deep feedforward neural
  networks.
\newblock In \emph{Proceedings of the thirteenth international conference on
  artificial intelligence and statistics}, pages 249--256, 2010.

\bibitem[He et~al.(2015)He, Zhang, Ren, and Sun]{he2015delving}
Kaiming He, Xiangyu Zhang, Shaoqing Ren, and Jian Sun.
\newblock Delving deep into rectifiers: Surpassing human-level performance on
  imagenet classification.
\newblock In \emph{Proceedings of the IEEE international conference on computer
  vision}, pages 1026--1034, 2015.

\bibitem[Spearman(1904)]{spearman1904proof}
Charles Spearman.
\newblock The proof and measurement of association between two things.
\newblock \emph{The American journal of psychology}, 15\penalty0 (1):\penalty0
  72--101, 1904.

\bibitem[Kendall(1948)]{kendall1948rank}
Maurice~George Kendall.
\newblock Rank correlation methods.
\newblock 1948.

\bibitem[Cohen et~al.(2014)Cohen, West, and Aiken]{cohen2014applied}
Patricia Cohen, Stephen~G West, and Leona~S Aiken.
\newblock \emph{Applied multiple regression/correlation analysis for the
  behavioral sciences}.
\newblock Psychology Press, 2014.

\end{thebibliography}

\end{document}